\documentclass{article}

\usepackage[preprint]{neurips_2023}

\usepackage[utf8]{inputenc} 
\usepackage[T1]{fontenc}    
\usepackage{hyperref}       
\usepackage{url}            
\usepackage{booktabs}       
\usepackage{amsfonts}       
\usepackage{amsmath,amssymb}
\usepackage{nicefrac}       
\usepackage{microtype}      
\usepackage{xcolor}         
\usepackage{wrapfig}
\usepackage{float}
\usepackage{graphicx}
\usepackage{colortbl}
\usepackage{multirow}
\usepackage{arydshln}
\usepackage{marvosym}
\usepackage{wasysym}
\usepackage{subcaption}
\usepackage{pifont}
\PassOptionsToPackage{numbers, compress}{natbib}

\title{
Mining Open Semantics from CLIP: A Relation Transition Perspective for Few-Shot Learning}

\author{%
    Cilin Yan \\
    Beihang University\\
    \texttt{clyan@buaa.edu.cn} \\
    \And
    Haochen Wang \\
    University of Amsterdam \\
    \texttt{h.wang3@uva.nl} \\
    \And
    Xiaolong Jiang, Yao Hu, Xu Tang\\
    Xiaohongshu\\
    \texttt{laige@xiaohongshu.com}
    \And
    Guoliang Kang\footnotemark[1]\\
    Beihang University\\
    \texttt{kgl.prml@gmail.com} \\
    \And
    Efstratios Gavves\\
    University of Amsterdam\\
    \texttt{egavves@uva.nl}
}

\begin{document}

\maketitle


\begin{abstract}
    
Contrastive Vision-Language Pre-training~(CLIP) demonstrates impressive zero-shot capability.
The key to improve the adaptation of CLIP to downstream task with few exemplars lies in 
how to effectively model and transfer the useful knowledge embedded in CLIP.
Previous work mines the knowledge typically based on the limited visual samples and close-set semantics (\emph{i.e.}, 
within target category set of downstream task). 
However, the aligned CLIP image/text encoders contain abundant relationships between visual features and almost 
\textit{infinite open semantics}, 
which may benefit the few-shot learning but remains unexplored. 
In this paper, we propose to mine open semantics as anchors to perform a relation transition from 
image-anchor relationship to image-target relationship to make predictions.
Specifically,  we adopt a transformer module which takes the visual feature as "Query",
the text features of the anchors as "Key" and the similarity matrix between the text features of 
anchor and target classes as "Value".
In this way, the output of such a transformer module represents the relationship between the image and target categories, 
\emph{i.e.}, the classification predictions.
To avoid manually selecting the open semantics, we make the [CLASS] token of input text embedding learnable. 
We conduct extensive experiments on eleven representative classification datasets.
The results show that our method performs favorably against previous state-of-the-arts considering few-shot classification settings. 

\end{abstract}
\section{Introduction}

Recent years have witnessed the rising of large-scale vision-language pre-trained models~\cite{radford2021learning,jia2021scaling,yao2021filip,li2022fine,lee2022uniclip,gao2022pyramidclip,zhou2022non}.
Among those visio-language models, CLIP~\cite{radford2021learning} is a widely-adopted representative due to its superior zero-shot capability on downstream tasks~\cite{wang2024learn}.
CLIP jointly trains the text and image encoders with image-text pairing supervision. 
Benefiting from the shared image and text feature space, CLIP can be directly adopted to recognize images from novel categories by
examining the degrees of alignment between image features and text features of novel categories.
Despite the amazing zero-shot performance of CLIP, there remains huge potential to improve the adaptation ability of CLIP if a few exemplars 
from the downstream visual task are accessible during training.

There are a few works that investigate how to improve the few-shot adaptation performance of CLIP~\cite{radford2021learning}. The key to improve the adaptation of CLIP lies in how to extract and transfer the useful knowledge of CLIP in terms of specific downstream tasks.
Most previous works extract the transferable knowledge only with limited visual samples and close-set definitions of categories (\emph{i.e.},
the category set of downstream task), including prompt tuning~\cite{zhou2022learning,zhou2022conditional}, classifier fine-tuning~\cite{he2022synthetic}, or image-image relation modeling~\cite{zhang2022tip}, \emph{etc.} However, as we know, CLIP is trained with large-scale open datasets and thus contains abundant relationships across \textit{almost infinite open semantics}. How to extract and utilize such semantic relation knowledge to improve the few-shot learning of CLIP remains unexplored.

\begin{wrapfigure}{r}{0.5\textwidth}
  \centering
  \includegraphics[width=1.0\linewidth]{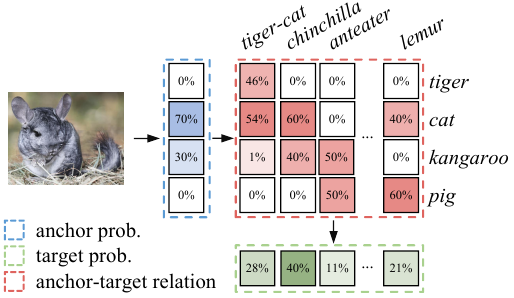}
  \vspace{-4mm}
\caption{
Transit Image-Anchor Relationship to Image-Target Relationship via Anchor-Target Relation Matrix.
}
\vspace{-5mm}
\label{fig:class_relation}
\end{wrapfigure}

As shown in Fig.~\ref{fig:class_relation}, we aim to recognize samples from target categories (green), \emph{i.e.}, "tiger-cat", "chinchilla", "anteater", and "lemur".
We manually select some other categories as anchors (blue), \emph{i.e.}, "tiger", "cat", "kangaroo", and "pig". 
In order to recognize the relation between image samples and target categories, we may first compute the similarities between 
image samples and the anchors to denote the image-anchor relation. 
We may also model the anchor-target relation by computing the similarities between anchors and target categories, as shown in Fig.~\ref{fig:class_relation}.
Through the transition of anchor-target relation matrix, the image-target relations can be obtained. 
A reasonable prior is that the image-target relations should keep consistent after relation transition, 
\emph{e.g.}, an image of "chinchilla" should be classified into the "chinchilla" category, no matter whether relation transition is performed.

In this paper, we propose to mine open semantics as anchors to perform a relation transition from 
image-anchor relation to image-target relation to facilitating few-shot learning.
Inspired by the illustration of Fig.~\ref{fig:class_relation},  we design the relation transition module (RTM) as a transformer 
decoder architecture which takes the visual feature as "Query",
the text features of the anchors as "Key" and the similarity matrix between the text features of 
anchors and target classes as "Value".
In this way, the output of RTM represents the relationship between the image and target categories, 
\emph{i.e.}, the classification predictions.
To avoid manually selecting the open semantics, we make the [CLASS] token of input text embedding learnable. 
During training, we freeze the CLIP encoders and 
impose cross-entropy loss with labeled visual samples to update the learnable [CLASS] token and the transformer module.
Via the transition of open semantics, we expect richer semantic relationships can be assembled to reduce the prediction error.
We name our framework as \textbf{R}elation \textbf{T}ransition with \textbf{O}pen \textbf{S}emantics (RTOS).
We conduct extensive experiments on eleven few-shot benchmarks and show that our method performs favorably against previous 
state-of-the-arts.

In a nutshell, our contributions can be summarized as follows
\begin{itemize}
    \item We propose a new perspective which aims to utilize the abundant semantic knowledge encoded in CLIP to benefit the few-shot learning task, \emph{i.e.}, mining open semantics as anchors to perform the relation transition.
    Via the transition of open semantics, we expect richer semantic relationships can be assembled to reduce the prediction error. 
    \item We propose to learn the open semantics via a learnable semantic token of text input and use a transformer module to perform relation transition from image-anchor relation to image-target relation. 
    With such designs, we avoid manually selecting anchors and make semantic relation modeling more adapted to the downstream task.
    \item Extensive experiments demonstrate our RTOS performs favorably against previous state-of-the-arts, \emph{e.g.}, 
    on Flowers102 datasets, we achieve 80.8\%, outperforming previous state-of-the-art method by 7.3\%. Moreover, ablations verify the effectiveness and necessity of each of our designs.
\end{itemize}


%
%
%

\section{Related Work}

\subsection{Traditional Few-shot Learning}

In the domain of traditional few-shot learning, researchers commonly curate multiple few-shot train and test sets, each comprising a limited number of training samples along with their corresponding test samples.
Earlier research on few-shot learning primarily focused on generative models that employed intricate iterative inference methodologies~\cite{fei2006one,lake2011one}. 
The success of deep learning-based methods in data-rich environments~\cite{krizhevsky2017imagenet,he2016deep,simonyan2014very} has led to a growing interest in adapting these approaches to handle few-shot learning scenarios.
To this end, various successful methods for few-shot learning have been proposed and leveraged techniques like meta-learning, metric learning, transfer learning, and transductive learning.
Meta-learning-based approaches~\cite{finn2017model,ravi2017optimization} involve training an auxiliary parameterization net that is capable of learning how to parameterize a feed-forward classification problem in terms of few-shot sample set.
Metric-learning-based approaches~\cite{bateni2020improved,snell2017prototypical,vinyals2016matching,sung2018learning} aim to learn a set of projection functions such that when represented in this embedding, images are easy to recognize using simple nearest neighbor or linear classifiers.
Transfer-learning-based methods~\cite{hariharan2017low,qi2018low} typically work by pre-training a deep neural network on a source dataset with abundant labeled examples, and fine-tuning the network on a few-shot learning task with limited labeled examples in the target dataset to improve performance.
Transductive-learning-based methods~\cite{dhillon2019baseline,joachims1999transductive,liu2018learning} typically work by leveraging the unlabeled examples in the target dataset to improve the few-shot learning performance, by exploiting the similarities and relationships between unlabeled and labeled examples.

\subsection{VLM-based Few-shot Learning}
Recently, AI has grown towards the dominant paradigm of learning foundation models~\cite{bommasani2021opportunities} from large-scale web data.
The development of Vision-Language Pre-trained Models~(VLM)~\cite{radford2021learning,jia2021scaling,yao2021filip,li2022fine,lee2022uniclip,gao2022pyramidclip,zhou2022non} has been particularly rapid, thanks to the low cost of collecting image-text pairs from the web. 
With the emergence of Vision-Language Models, a new few-shot evaluation protocol has been developed, as implemented by recent works on few-shot adaptation with CLIP~\cite{radford2021learning}.
In this new protocol, the meta-training phase is replaced with pre-trained CLIP models, and the official test splits of each dataset are utilized as the test sets.
Prompt tuning~\cite{zhou2022learning,zhou2022conditional} and adapter~\cite{zhang2022tip,gao2021clip} have emerged as representative parameter-efficient learning paradigms that adapt VLM to downstream tasks.
Prompt tuning of CLIP~\cite{radford2021learning,zhou2022conditional,lu2022prompt,xing2022class,zhu2022prompt,sun2022dualcoop} draws inspiration from the successful implementation of prefix-tuning in language models~\cite{deng2022rlprompt,gao2020making,haviv2021bertese,jiang2020can}, specifically aiming to extract broader text features.
Similarly, CLIP-Adapter~\cite{gao2021clip} and Tip-Adapter~\cite{zhang2022tip} draw inspiration from parameter-efficient fine-tuning methods~\cite{houlsby2019parameter,jia2022visual,zhang2020side} that optimize lightweight MLPs while keeping the encoder frozen. 
These approaches mine the knowledge typically based on the limited visual samples and close-set semantics. 
However, the aligned CLIP image/text encoders contain abundant relationships between visual features and almost infinite open semantics, 
which may benefit the few-shot learning but remains unexplored. 
In this paper, we focus on mining open semantics as anchors to perform a relation transition from 
image-anchor relationship to image-target relationship to make predictions.
\section{Method}
\label{method}

\begin{figure*}
\begin{center}
\includegraphics[width=1.0\linewidth]{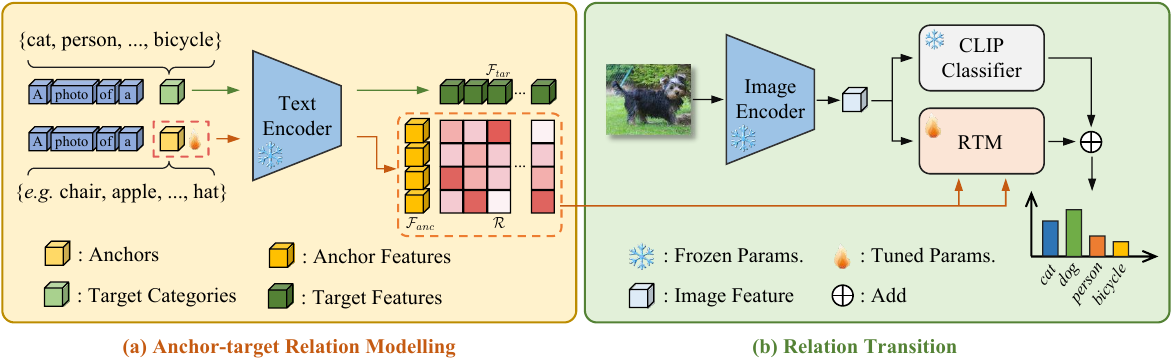}
\end{center}
   \vspace{-3mm}
\caption{\textbf{The Framework of RTOS.} 
\textcolor[rgb]{0.796875, 0.3984375, 0.0}{{(a) Anchor-target Relation Modelling}}: Input with target categories and anchors, CLIP text encoder is adopted to extract target features $\mathcal{F}_{tar}$ and anchor features $\mathcal{F}_{pro}$, which are used to model anchor-target relation matrix $\mathcal{R}$ between target categories and anchors. 
\textcolor[rgb]{0.35546875, 0.5546875, 0.1953125}{{(b) Relation Transition}}: Input a test image $\mathcal{I}$, CLIP image encoder is adopted to extract the image visual feature. Next, the Relation Transition Module (RTM) $\Phi$ is applied to transition image-anchor relation to image-target relation. CLIP classifier is initialized using target features $\mathcal{F}_{tar}$. The knowledge acquired from relation transition is then integrated with CLIP's pre-trained knowledge, enabling accurate prediction.
}
\vspace{-4mm}
\label{fig:architecture}
\end{figure*}

The overall framework of \textbf{R}elation \textbf{T}ransition with \textbf{O}pen \textbf{S}emantics~(RTOS) we proposed is shown in Fig.~\ref{fig:architecture}.
Given an image to recognize, we obtain its feature through CLIP's image encoder. 
We construct two groups of text inputs. 
One group corresponds to the target classes, which consists of manually designed prompts and different target class names; the other one corresponds to the anchors, which consists of manually designed prompts and different [CLASS] tokens. We make the [CLASS] token of anchor text input learnable to mine open semantics from CLIP.
Then the anchor-target relation matrix is constructed by computing similarities between anchor text features and target text features (Sec.~\ref{sec:icr_model}). 
In our framework, we design a relation transition module (RTM) based on the transformer architecture which takes image features as Query, anchor text features as Key, and anchor-target relation matrix as Value. 
The output of RTM denotes the relations between image and target classes. 
Finally, we combine the output of RTM with the output of CLIP's zero-shot classifier to produce the final prediction (Sec.~\ref{relation_based_cls}).


\subsection{Anchor-target Relation Modelling}
\label{sec:icr_model}


In this paper, our primary task focuses on zero-shot and few-shot image classification, where we aim to classify images into their respective categories with limited or no training examples available for the target categories. 
The target categories serve as a set of labels for the classification task. 
Our primary goal is to mine these open semantics to improve the CLIP's classification performance.
We use anchors as cues derived from the open semantics within CLIP. 
Anchors serve as a bridge to connect and transition image features to their corresponding target categories in image classification task.
To be more specific, anchors can take different forms depending on their origin and semantics. 
For instance, anchors can be category names sampled from cross-dataset category list, which carry explicit semantic information. 
Alternatively, they may be randomly initialized class tokens that do not hold any clear semantic information.
We model the anchor-target relation, which is used to transition the image-anchor relation to image-target relation.

The anchor-target relation is modeled using the CLIP text encoder.
Following~\cite{zhou2022learning}, we define the target categories prompts $\mathcal{T}_{tar}$ and the anchors prompts $\mathcal{T}_{anc}$ given to the text encoder as follows:
\begin{align}
\mathcal{T}_{tar,i}\;&=[\boldsymbol{v}_1, \boldsymbol{v}_2, ...,\boldsymbol{v}_M, \boldsymbol{s}_{tar, i}\;] \\
\mathcal{T}_{anc,j}&=[\boldsymbol{v}_1, \boldsymbol{v}_2, ...,\boldsymbol{v}_M, \boldsymbol{s}_{anc, j}]
\end{align}
where $i\in \{1, 2, ..., C_{tar} \}$ is the target categories index, $C_{tar}$ denotes the number of target categories, $\boldsymbol{s}_{tar, i}$ denotes word embedding of the $i$-th target categories name $\mathrm{s}_{i}$,
$j \in \{1, 2, ..., C_{anc} \}$ is the anchors index, $C_{anc}$ denotes the number of anchors, the $\boldsymbol{s}_{anc, j}$ denotes word embedding of the $j$-th anchors name $\mathrm{s}_{j}$.
The anchors are initialized by cross-dataset category list or randomly initialized in the experiments. 
$[\boldsymbol{v}_1, \boldsymbol{v}_2, ...,\boldsymbol{v}_M]$ denotes word embedding of the prompt sentences prefix (\textit{e.g.} $\texttt{"A photo of \{\}."}$).

After obtaining the text feature, as shown in Fig.~\ref{fig:architecture} (a), by forwarding the target categories prompts $\mathcal{T}_{tar}$ and the anchors prompts $\mathcal{T}_{anc}$ to the CLIP text encoder, we can obtain the target L2 normalized features $\mathcal{F}_{tar} \in \mathbb{R}^{C_{tar} \times D}$ and the anchor L2 normalized features $\mathcal{F}_{anc} \in \mathbb{R}^{C_{anc} \times D}$. 
Where $D$ denotes the feature dimension of the CLIP's visual-language feature space
(\textit{e.g.} $D=512$ for ResNet50~\cite{he2016deep} backbone in CLIP). 
The features of target categories and anchors are used to build anchor-target relation matrix $\mathcal{R} \in \mathbb{R}^{ C_{tar} \times C_{anc} }$ to model the anchor-target relation. 
The anchor-target relation matrix $\mathcal{R}$ is defined in Eq.~\ref{cls_shift_matrix}.

\begin{equation}
     \mathcal{R}_{i,j} = \frac{\exp(\cos( \mathcal{F}_{tar, i},\mathcal{F}_{anc, j} )/\tau)}{
\sum\nolimits_{j=1}^{C_{anc}} \exp(\cos( \mathcal{F}_{tar, i},\mathcal{F}_{anc, j} )/\tau) 
} 
\label{cls_shift_matrix}
\end{equation}

\subsection{Relation Transition}
\label{relation_based_cls}

%

Upon establishing anchor-target relation, we employ the Relation Transition Module (RTM) $\Phi$ to transform the image-anchor relation into the image-target relation for classification prediction. This process essentially involves mining the open semantics embedded within the CLIP framework, which in turn ensures better representation and recognition of the target categories.
As shown in Fig.~\ref{fig:architecture} (b), by forwarding the test image $\mathcal{I} \in \mathbb{R} ^ {H \times W \times 3}$ to the CLIP image encoder, we can obtain the test image L2 normalized feature $f_{test} \in \mathbb{R} ^ {D}$.
The Relation Transition Module (RTM) $\Phi$ is a cross-attention module, which accepts the image visual feature $f_{test}$ as "Query", anchor features $\mathcal{F}_{anc}$ as "Key", and the anchor-target relation matrix $\mathcal{R}$ as "Value" to transition image-anchor relation $\mathcal{P}_{anc} \in \mathbb{R} ^ {C_{anc}}$ to image-target relation $\mathcal{P}_{tar} {\color{red} \in \mathbb{R}^{C_{tar}}}$, defined as $\mathcal{P}_{tar} = \Phi (f_{test}, \mathcal{F}_{anc}, \mathcal{R})$. 

Based on the reasonable prior that the image-target relations remain consistent through the relation transition process, it is possible to perform image classification prediction in the zero-shot setting by utilizing the relationships between images, target categories, and anchors. To accomplish this, we first map the images and target categories onto the anchor space. Next, we determine the image-target relations within this mapped space. Based on this assumption, the anchor-target relation $\mathcal{R}$ acts as the mapping for target categories onto anchors. Similarly, we can define the mapping of an image onto anchors through Eq.~\ref{eq:p_anc}.

\begin{equation}
    \mathcal{P}_{anc,j} = \frac{
    \exp( \cos(f_{test}, \mathcal{F}_{anc,j}) / \tau )
    }{
    \sum_{j=1}^{C_{anc}} \exp( \cos(f_{test}, \mathcal{F}_{anc,j}) / \tau )
    } 
    \label{eq:p_anc}
\end{equation}

Then, the Relation Transition Module (RTM) performs a relation transition from  image-anchor relation $\mathcal{P}_{anc}$ to image-target relation $\mathcal{P}_{tar}$. This transition is defined as 


\begin{equation}
    \mathcal{P}_{tar,i} = \frac{
    \exp( \cos(\mathcal{P}_{anc}, \mathcal{R}_{j}) / \tau' )
    }{
    \sum_{i=1}^{C_{tar}} \exp( \cos(\mathcal{P}_{anc}, \mathcal{R}_{i}) / \tau' )
    } .
    \label{eq:p_tar_tr}
\end{equation}

By following this approach, we are able to make use of the open semantics within the pre-trained CLIP framework, which allows for better representation and recognition of the target categories, ultimately leading to improved performance in image classification tasks.

When we have a limited amount of data available for training, it is possible that the manually designed relation transitions are not optimal. Therefore, we employ the Transformer model, allowing the network to learn more effective relation transition strategies autonomously. We use the transformer architecture as RTM $\Phi$. The image-target relation is obtained by

\begin{equation}
    \mathcal{P}_{tar} = \text{Transformer} (f_{test}, \mathcal{F}_{anc}, \mathcal{R}).
\end{equation}

Then, the relation-based prediction $\mathcal{P}_{tar}$ works in conjunction with CLIP's zero-shot prediction $\mathcal{P}_{clip}$ to produce the final prediction $\mathcal{P}_{final}$.

\begin{equation}
    \mathcal{P}_{final} = \mathcal{P}_{clip} + \alpha \mathcal{P}_{tar}
\end{equation}

Using this method, we harness anchors as open semantics, which enable the effective relation transition from image-anchor relation to image-target relation.

\subsection{Training and Inference}
\label{train_and_inference}

\textbf{Zero-shot Setting.}
In the zero-shot setting, our method does not require any training, and we refer to it as ZS-RTOS.
First, we manually select anchors from cross-dataset category list. 
Then, we build the anchor-target relation matrix $\mathcal{R}$ to model anchor-target relationship. 
The anchor-target relation matrix only needs to be created once, which means that we only need to perform one forward pass of the CLIP text encoder. Once the anchor-target relation is established, the CLIP text encoder can be discarded.
Next, the Relation Transition Module $\Phi$ is applied to convert the image-anchor relation $\mathcal{P}_{anc}$ to the image-target relation $\mathcal{P}_{tar}$. 
Finally, the relation-based prediction $\mathcal{P}_{tar}$ collaborates with CLIP's zero-shot prediction $\mathcal{P}_{zs}$ to produce the final prediction $\mathcal{P}_{final}$.

\textbf{Few-shot Setting.}
ZS-RTOS offers a substantial boost to CLIP by incorporating fresh open semantics relation transition knowledge. 
Moreover, RTOS remains compatible with situations that involve training data. 
In the few-shot scenario, a limited amount of data can aid RTOS in acquiring superior anchors, thereby elevating its performance even further. 
RTOS focuses on updating the anchor embeddings (which can be pre-designed or randomly initialized) to learn anchor classes that provide significant performance improvements to CLIP. 
For few-shot setting, we use cross entropy loss to update RTOS.
With the updates made to RTOS, it attains state-of-the-art performance on 11 widely adopted datasets. 

In addition to the relation transition method based on the consistency prior described above, we also explored a relation transition approach based on the total probability formula. 
In this approach, the anchor-target relationship is defined in Eq.~\ref{cls_shift_matrix_p}, and the relation transition can be expressed as $\mathcal{P}_{tar} = \mathcal{R}\mathcal{P}_{anc}$. 

\begin{equation}
     \mathcal{R}_{i,j} = \frac{\exp(\cos( \mathcal{F}_{tar, i},\mathcal{F}_{anc, j} )/\tau)}{
\sum\nolimits_{i=1}^{C_{tar}} \exp(\cos( \mathcal{F}_{tar, i},\mathcal{F}_{anc, j} )/\tau) 
} 
\label{cls_shift_matrix_p}
\end{equation}

Moreover, we also investigated the image-image relation method. In this approach, we use the training image features as the input keys for the Relation Transition Model (RTM), and the one-hot labels of the training images serve as the RTM's values. The performance of the methods based on the total probability formula and image-image relation is evaluated in Experiment Section.

\section{Experiments}

\begin{table*}
\begin{center}
\scalebox{1.0}{
\begin{tabular}{lcccccccccccc}
\toprule[1pt]
       \multicolumn{1}{l}{Method} & \rotatebox{90}{Average} & \rotatebox{90}{EuroSAT} & \rotatebox{90}{Caltech101} &  \rotatebox{90}{DTD} & \rotatebox{90}{UCF101} & \rotatebox{90}{StanfordCars} & \rotatebox{90}{OxfordPets} & \rotatebox{90}{SUN397} & \rotatebox{90}{Flowers102} & \rotatebox{90}{FGVCAircraft} & \rotatebox{90}{ImageNet} & \rotatebox{90}{Food101} \\
\midrule
         CLIP &                     58.9 &    37.5 &       85.9 & 42.2 &   61.4 &         55.7 &       85.8 &   58.5 &       66.0 &         17.1 &     60.3 &    77.3 \\ 
\rowcolor[gray]{.9}
 Random  &                     59.0 &    36.2 &       86.1 & 42.9 &   61.7 &         55.9 &       85.9 &   \textbf{58.9} &       66.2 &         17.2 &     \textbf{60.4} &    77.3 \\
\rowcolor[gray]{.9}
Selected &                     \textbf{59.4} &    \textbf{39.3} &       \textbf{87.5} & \textbf{43.3} &   \textbf{62.0} &         \textbf{55.8} &       \textbf{85.8} &   \textbf{58.9} &       \textbf{66.1} &         \textbf{17.2} &     \textbf{60.4} &    \textbf{77.4} \\
\bottomrule[1pt]
\end{tabular}
}
\vspace{-1pt}
\caption{\textbf{Accuracy of zero-shot learning on the 11 datasets.} 
The "Average" denotes the average accuracy over the 11 datasets. 
The "Random" denotes the accuracy of method with randomly initialized anchor class embeddings. 
The "Selected" denotes the accuracy of method with manually selected anchors. The class names of anchors are chosen from the class definitions in the other 10 datasets.
\vspace{-4mm}}
\label{tab:comparison_on_zeroshot}
\end{center}
\end{table*}

\begin{table*}
\begin{center}
\scalebox{1.0}{
\begin{tabular}{lccccc}
\toprule[1pt]
\multicolumn{1}{c}{\multirow{2}{*}{Method}} & \multicolumn{5}{c}{Number of shots} \\
\cmidrule(lr){2-6}
                        & 1     & 2     & 4    & 8    & 16   \\
\cmidrule(lr){1-6}
\ding{71} Linear probe CLIP~\cite{zhou2022learning} & 36.7 & 47.6 & 57.2 & 65.0 & 71.1 \\
\ding{71}              CoOp~\cite{zhou2022learning} & 59.6 & 62.3 & 66.8 & 69.9 & 73.4 \\
\ding{71}      CLIP-Adapter~\cite{gao2021clip} & 62.7 & 65.5 & 68.6 & 71.3 & 74.4 \\
\ding{71}           ProGrad~\cite{zhu2022prompt} & 62.6 & 64.9 & 68.5 & 71.4 & 73.9 \\
\ding{71}       Tip-Adapter~\cite{zhang2022tip} & 62.3 & 64.6 & 66.5 & 68.5 & 70.3 \\
\ding{71}     Tip-Adapter-F~\cite{zhang2022tip} & 64.6 & 66.7 & 69.7 & 72.4 & 75.8 \\
\rowcolor[gray]{.9}
\ding{71}    RTOS & 65.5 & 67.4 & 70.3 & 72.7 & 75.8 \\ 
\rowcolor[gray]{.9}
\ding{71}   RTOS-IR & \textbf{66.1} & \textbf{68.2} & \textbf{71.1} & \textbf{73.6} & \textbf{76.6} \\
\cmidrule(lr){1-6}
\ding{70}    Synthetic~\cite{he2022synthetic} & 60.5 & 62.9 & 67.0 & 69.9 & 73.1 \\ 
\rowcolor[gray]{.9}
\ding{70}    RTOS & 62.4 & 64.0 & 66.9 & 69.2 & 72.5 \\ 
\rowcolor[gray]{.9}
\ding{70}   RTOS-IR & \textbf{62.7} & \textbf{64.6} & \textbf{67.6} & \textbf{70.2} & \textbf{73.5} \\
\bottomrule[1pt]
\end{tabular}
}
\vspace{-1pt}
\caption{\textbf{Comparison with previous few-shot methods.} \ding{71} denotes the average accuracy of 11 datasets, \ding{70} denotes the average accuracy of 8 datasets reported in~\cite{he2022synthetic}. 
Our methods outperform existing state-of-the-art methods in few-shot classification settings.
\vspace{-5mm}}
\label{tab:comparison_on_fewshot}
\end{center}
\end{table*}

\subsection{Experimental Setups}

\textbf{Datasets}
Following CoOp~\cite{zhou2022learning}, we conduct experiments for RTOS on 11 widely-used image classification datasets: ImageNet~\cite{deng2009imagenet}, StandfordCars~\cite{krause20133d}, UCF101~\cite{soomro2012ucf101}, Caltech101~\cite{fei2004learning}, Flowers102~\cite{nilsback2008automated}, SUN397~\cite{xiao2010sun}, DTD~\cite{cimpoi2014describing}, EuroSAT~\cite{helber2019eurosat}, FGVCAircraft~\cite{maji2013fine},  OxfordPets~\cite{parkhi2012cats}, and Food101~\cite{bossard2014food}. 
These datasets constitute a comprehensive benchmark, which covers a diverse set of vision tasks including the classification of generic objects, scenes, actions, and fine-grained categories, as well as specialized tasks like recognizing textures and satellite imagery.

\textbf{Implementation Details}
We follow the data preprocessing protocol in CLIP~\cite{radford2021learning}, which is composed of random cropping, resizing, and random horizontal flip.
Following Tip-Adapter~\cite{zhang2022tip}, we adopt prompt ensembling for experiments on ImageNet and use single handcrafted prompt on the other 10 datasets. 
For the CLIP~\cite{radford2021learning} backbone, we utilize ResNet-50~\cite{he2016deep} as the visual encoder.
We obtain the pre-trained weights of both encoders from~\cite{radford2021learning} and freeze them during training.
The batch size is set to 256. We adopt the AdamW~\cite{kingma2014adam} optimizer with learning rate set to 0.00001 and a cosine scheduler. 
Following Tip-Adapter~\cite{zhang2022tip}, we train 100 epochs on the EuroSAT dataset and 20 epochs on the other 10 datasets. 
We set $\tau$ and $\tau'$ to 0.01. We set hype-parameter $\alpha$ following Tip-Adapter~\cite{zhang2022tip}.
For the zero-shot setting, we directly test the model's performance on the full test set. 
Besides, following~\cite{radford2021learning}, we consider the 1-, 2-, 4-, 8-, 16-shot settings, where we utilize 1, 2, 4, 8, 16 labeled samples to train the model and then evaluate the trained model on the full test set.

\subsection{Comparison under Zero-Shot Setting}

We conduct experiments with different anchor selection ways under zero-shot setting, and compare them to the zero-shot CLIP baseline in Tab.~\ref{tab:comparison_on_zeroshot}. 
The average accuracy across the 11 datasets is reported for comparison. 
As shown in Tab.~\ref{tab:comparison_on_zeroshot}, without any training and extra annotations, selecting anchors to perform relation transition performs favorably against the zero-shot CLIP baseline.
Specifically, the method with manually selected anchors outperforms CLIP by 0.5\%. 
Notably, on the EuroSAT and Caltech101 datasets, the zero-shot accuracy of ours outperforms CLIP by 1.8\% and 1.6\% respectively. 
These improvements justify that with carefully selected anchors, performing the relation transition benefits the adaption of CLIP to downstream tasks.
However, method with randomly selected anchors barely yields accuracy improvement, which inspires us to determine the effective anchors in a learnable way.


\begin{table*}
\begin{center}
\scalebox{1.0}{
\begin{tabular}{lcccccccc}
\toprule[1pt]
\multicolumn{1}{c}{\multirow{2}{*}{Alias}} & \multicolumn{3}{c}{Knowledge} & \multicolumn{5}{c}{Number of shots} \\
\cmidrule(lr){2-4} \cmidrule(lr){5-9}
\multicolumn{1}{c}{} & Cons. & Prob. & Image. & 1 & 2 & 4 & 8 & 16 \\
\cmidrule(lr){1-9}
RTOS-C & \checkmark &  &  & 65.0 & 67.0 & 70.0 & 72.4 & 75.4 \\
RTOS-P  &  & \checkmark &  & 63.3 & 63.9 & 65.2 & 67.4 & 71.6 \\
\rowcolor[gray]{.9}
RTOS & \checkmark & \checkmark &  & 65.5 & 67.4 & 70.3 & 72.7 & 75.8 \\
\multicolumn{1}{c}{-} & \checkmark &  & \checkmark & 65.9 & 68.0 & 70.9 & 73.5 & \textbf{76.6} \\
\rowcolor[gray]{.9}
RTOS-IR & \checkmark & \checkmark & \checkmark & \textbf{66.1} & \textbf{68.2}& \textbf{71.1}   & \textbf{73.6} & \textbf{76.6} \\
\bottomrule[1pt]
\end{tabular}
}
\vspace{-1pt}
\caption{\textbf{Different configuration of our methods.} 
Average accuracy of the 11 datasets. 
Cons. denotes the approach based on consistency prior, 
Prob. denotes the approach based on the total probability formula. 
Image. denotes the approach based on image-image relation. The methods highlighted in gray represent the approaches we compare in the experimental Sec.~\ref{sec:cpr_with_fs_method}.
\vspace{-5mm}}
\label{tab:comparison_Knowledge}
\end{center}
\end{table*}

\subsection{Comparison with SOTA few-shot methods}
\label{sec:cpr_with_fs_method}

\begin{figure*}
\begin{center}
\includegraphics[width=1.0\linewidth]{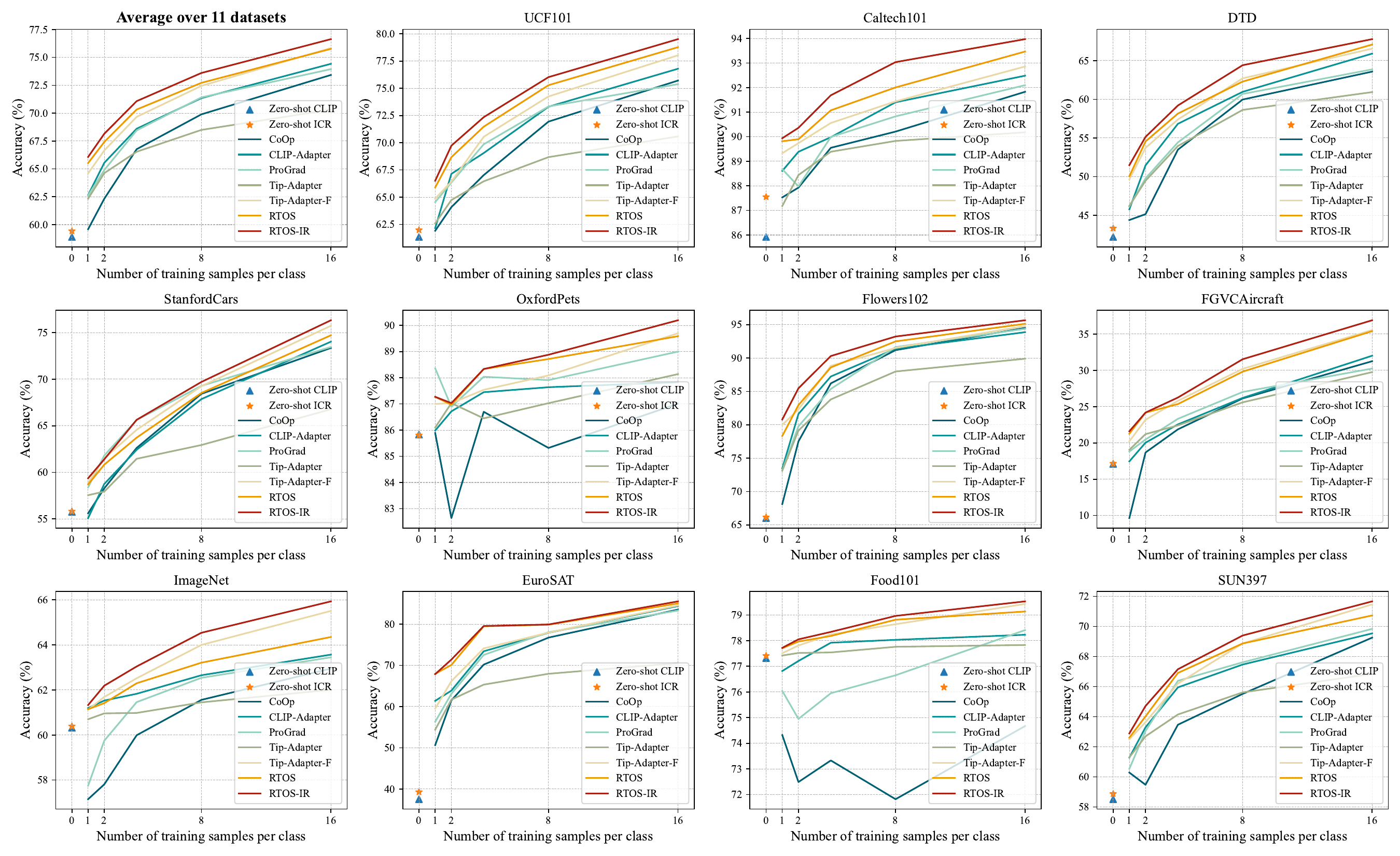}
\end{center}
   \vspace{-3mm}
\caption{\textbf{Accuracy of zero-shot and few-shot learning on the 11 datasets.} 
Overall, Zero-shot RTOS improves Zero-shot CLIP without any data. 
RTOS consistently surpasses all previous start-of-the-art methods by efficiently fine-tuning the anchor class embeddings.
}
\vspace{-4mm}
\label{fig:acc_all_curve}
\end{figure*}

We compare our method to previous state-of-the-art methods including Zero-shot CLIP~\cite{radford2021learning}, Linear-probe CLIP~\cite{radford2021learning}, CoOp~\cite{zhou2022learning}, CLIP-Adapter~\cite{gao2021clip}, ProGrad~\cite{zhu2022prompt}, Tip-Adapter~\cite{zhang2022tip}, and Synthetic~\cite{he2022synthetic} in Tab.~\ref{tab:comparison_on_fewshot} and Fig.~\ref{fig:acc_all_curve}. 
With several annotated images to train the [Class] tokens of anchors, the RTOS achieves new state-of-the-arts on average.
In particular, on the EuroSAT dataset, RTOS surpasses Tip-Adapter-F by 8.4\% and 3.9\% in 1-shot and 2-shot settings.
After integrating with the image-image relations (see Tab.~\ref{sec:cpr_with_fs_method}), the RTOS-IR outperforms all existing methods in all few-shot settings. 
For example, RTOS-IR obviously outperforms Tip-Adapter-F by 1.2\% and 0.8\% in 8-shot and 16-shot settings.
The results demonstrate that mining open semantic relations from CLIP greatly benefits few-shot learning, which justifies the design of our framework.

\subsection{Ablation Studies}





\begin{figure*}
\begin{center}
\includegraphics[width=0.819\linewidth]{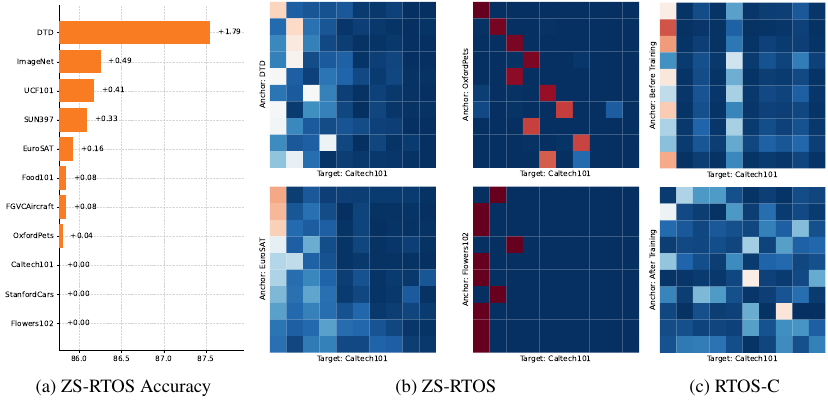}
\end{center}
   \vspace{-3mm}
\caption{
\textbf{Ablation on the relation between target categories and anchors.} 
  (a) illustrates the performance gain in accuracy of CLIP on the Caltech101 dataset when using different datasets' categories as anchors. 
  (b) show association heatmaps between different anchors and target categories.
  (c) show association heatmaps before and after training.
  \vspace{-5mm}
}
\vspace{-4mm}
\label{fig:heatmap}
\end{figure*}



\textbf{Effect of different choices of anchors.}
In this section, we investigate which type of anchor will benefit the adaptation of CLIP. 
We conduct experiments with our ZS-RTOS framework on the Caltech101~\cite{fei2004learning} dataset. 
We use all 11 datasets' categories as anchors and evaluate corresponding performance gains.
As shown in Fig.~\ref{fig:heatmap}~(a), different instantiations of anchors yield different zero-shot accuracy.
For example, using categories from DTD~\cite{cimpoi2014describing} dataset as anchors achieves 1.79\% improvement compared to the zero-shot CLIP baseline, while using categories from Caltech101 itself or from Flowers102 dataset as anchors don't bring any performance improvement.

We also examine the anchor-target relation matrix in Fig.~\ref{fig:heatmap}~(b) to investigate the powerful pattern which may benefit the relation transition and final prediction. 
As we observed, if the marginal distribution of target classes (which can be obtained by averaging all the rows of the anchor-target relation matrix) is more balanced, the performance gain will be larger. 
For example, the marginal distribution of target classes is more balanced for DTD (1.79\% gain) than for Flowers102 (no gain).
Moreover, as shown in Fig.~\ref{fig:heatmap}~(c), the relation matrix between learned anchors and target classes exhibits similar pattern to that using DTD categories as anchors, verifying that our learnable way mines beneficial open semantics to facilitate effective relation transition.

\begin{table*}
\begin{center}
\scalebox{0.99}{
\begin{tabular}{llcccccc}
\toprule[1pt]
\multicolumn{8}{c}{Ablation Studies on RTOS-C}                                                                                                                                                               \\ \cmidrule(lr){1-8}
\multicolumn{2}{c}{shots}   & 0                        & 1                        & 2                        & 4                        & 8                        & 16                       \\ \cmidrule(lr){1-8}
                                & \multicolumn{1}{c}{human-designed}     & \multicolumn{1}{c}{\textbf{87.5}} & \multicolumn{1}{c}{89.6} & \multicolumn{1}{c}{\textbf{90.3}} & \multicolumn{1}{c}{90.9} & \multicolumn{1}{c}{92.1} & \multicolumn{1}{c}{92.4} \\ 
\multirow{-2}{*}{(a) Init. Method}  & \multicolumn{1}{c}{ \cellcolor[gray]{.9} rand. initialized} & 86.1                     & \textbf{89.8}                     & \textbf{90.3}                     & \textbf{91.3}                     & \textbf{92.9}                     & \textbf{93.8}                     \\ \cmidrule(lr){1-8}
                                & \multicolumn{1}{c}{text features}    & -                        & 89.6                     & \textbf{90.4}                     & 91.0                     & 92.2                     & 92.6                     \\
\multirow{-2}{*}{(b) Tuned Params.} & \multicolumn{1}{c}{ \cellcolor[gray]{.9} class embeddings} & -                        & \textbf{89.8}                     & 90.3                     & \textbf{91.3}                     & \textbf{92.9}                     & \textbf{93.8}    \\

\cmidrule(lr){1-8}
& \multicolumn{1}{c}{direct}    & - 
& 88.4  & 89.1   &  89.8  &  90.9  &  91.6 \\
\multirow{-2}{*}{(c) Relation Transition Module} & \multicolumn{1}{c}{ \cellcolor[gray]{.9} transformer} & - &
\textbf{89.8} & \textbf{90.3} & \textbf{91.3} & \textbf{92.9} & \textbf{93.8}    \\
\bottomrule[1pt]
\end{tabular}
}
\vspace{-1pt}
\caption{
\textbf{Classification Accuracy on Caltect101.} 
(a) Ablation studies on different initialization methods. 
(b) Ablation studies on different finetuned parameters.
(c) Ablation studies on directly using Eq.~\ref{eq:p_anc} and Eq.~\ref{eq:p_tar_tr} or using transformer as relation transition modele.
The methods marked in gray indicate the configurations adopted by the RTOS.
\vspace{-5mm}}
\label{tab:ablation_init_method}
\end{center}
\end{table*}

\textbf{Effect of different initialization ways of anchors.}
Tab.~\ref{tab:ablation_init_method}~(a) shows the zero-shot~(0) and few-shot~(1-16) performance on Caltect101 dataset under two different anchor class initialization methods.
The human-designed method achieves 87.5\% under the zero-shot setting, outperforming the random initialization method by 0.6\%.
On the contrary, under different few-shot settings, the random initialization method outperforms the human-designed method.
That is because the random initialization of anchor embeddings allows for more flexible learning of task-beneficial semantics from available data. On the contrary, using manually designed categories as anchors may restrict such learning and adaptation processes and thus achieve sub-optimal results.


\textbf{Effect of different parameters to tune.}
Tab.~\ref{tab:ablation_init_method}~(b) presents the classification accuracy on the Caltect101 dataset for various finetuned parameters. 
The results indicate that fine-tuning anchor class embeddings is a more effective method compared to directly fine-tuning text features. 
Directly fine-tuning text features may have a risk of overfitting to the training data.

\textbf{Effect of learnable relation transition.} 
The results in Tab.~\ref{tab:ablation_init_method}~(c) demonstrate that employing a transformer to learn different relation transitions outperforms the direct relation transition via multiplication between image-anchor relation vector and anchor-target relation matrix, showing the learnable way can better depict the relation transition process. 

\begin{table*}
\begin{center}
\scalebox{1.00}{
\begin{tabular}{cccccccccc}
\toprule[1pt]
\multicolumn{1}{c}{\multirow{2}{*}{Shots}} & \multicolumn{9}{c}{Number of Anchors} \\ \cmidrule(lr){2-10}
\multicolumn{1}{c}{} & 5 & 10 & 20 & 40 & 60 & 80 & 100 & 150 & 200 \\  \cmidrule(lr){1-10}
1 & 88.2 & 89.3 & 88.9 & 88.9 & 89.0 & \textbf{89.6} & 88.7 & 88.7 & 88.5 \\
2 & 89.3 & 89.2 & 89.5 & 89.3 & 89.5 & \textbf{89.7} & 89.4 & 89.3 & \textbf{89.7} \\
4 & 90.4 & 90.8 & 90.1 & 90.1 & 90.4 & \textbf{91.0} & 90.8 & 90.8 & 90.6 \\
8 & 90.2 & 90.8 & 91.0 & 91.4 & 91.2 & \textbf{92.0} & 91.5 & 91.9 & 91.4 \\
16 & 90.9 & 92.0 & 92.7 & 92.8 & 92.6 & 92.9 & 93.0 & 92.8 & \textbf{93.1} \\
\bottomrule[1pt]
\end{tabular}
}
\vspace{-1pt}
\caption{\textbf{Ablation studies on the number of anchors on Caltech101 dataset.}  
\vspace{-3mm}
}
\label{tab:ablation_aux_cls_number}
\end{center}
\end{table*}

\textbf{Effect of anchor numbers.}
In this section, we investigate the influence of the number of anchor classes over the few-shot performance of RTOS-C. 
As shown in Tab.~\ref{tab:ablation_aux_cls_number}, when the number of anchor categories is smaller than the number of target categories, RTOS-C brings negligible improvements. 
When the number of anchor categories is set to 80, RTOS-C achieves the best performance.



\section{Conclusions}

In this paper, we aim to utilize the abundant semantic knowledge encoded in CLIP to benefit the few-shot learning task by mining open semantics as anchors to perform the relation transition.
To this end, we propose RTOS, which learns the open semantics via a learnable semantic token of text input and uses a transformer module to perform relation transition from image-anchor relation to image-target relation.
Extensive experiments verify the effectiveness of our proposed method.

\textbf{Broader Impact and Limitations}
Our method will not introduce bias but it may be impacted by the bias contained in CLIP. 
Our method mines semantic knowledge from CLIP to benefit the few-shot learning. There still remain other useful priors which may benefit the CLIP-based few-shot learning to be investigated in the future.  



{\small
\bibliographystyle{ieeetr}
\bibliography{reference}

\begin{thebibliography}{10}

\bibitem{radford2021learning}
A.~Radford, J.~W. Kim, C.~Hallacy, A.~Ramesh, G.~Goh, S.~Agarwal, G.~Sastry,
  A.~Askell, P.~Mishkin, J.~Clark, {\em et~al.}, ``Learning transferable visual
  models from natural language supervision,'' in {\em International conference
  on machine learning}, pp.~8748--8763, PMLR, 2021.

\bibitem{jia2021scaling}
C.~Jia, Y.~Yang, Y.~Xia, Y.-T. Chen, Z.~Parekh, H.~Pham, Q.~Le, Y.-H. Sung,
  Z.~Li, and T.~Duerig, ``Scaling up visual and vision-language representation
  learning with noisy text supervision,'' in {\em International Conference on
  Machine Learning}, pp.~4904--4916, PMLR, 2021.

\bibitem{yao2021filip}
L.~Yao, R.~Huang, L.~Hou, G.~Lu, M.~Niu, H.~Xu, X.~Liang, Z.~Li, X.~Jiang, and
  C.~Xu, ``Filip: fine-grained interactive language-image pre-training,'' {\em
  arXiv preprint arXiv:2111.07783}, 2021.

\bibitem{li2022fine}
J.~Li, X.~He, L.~Wei, L.~Qian, L.~Zhu, L.~Xie, Y.~Zhuang, Q.~Tian, and S.~Tang,
  ``Fine-grained semantically aligned vision-language pre-training,'' {\em
  Advances in Neural Information Processing Systems}, 2022.

\bibitem{lee2022uniclip}
J.~Lee, J.~Kim, H.~Shon, B.~Kim, S.~H. Kim, H.~Lee, and J.~Kim, ``Uniclip:
  Unified framework for contrastive language-image pre-training,'' {\em
  Advances in Neural Information Processing Systems}, 2022.

\bibitem{gao2022pyramidclip}
Y.~Gao, J.~Liu, Z.~Xu, J.~Zhang, K.~Li, R.~Ji, and C.~Shen, ``Pyramidclip:
  Hierarchical feature alignment for vision-language model pretraining,'' {\em
  Advances in Neural Information Processing Systems}, vol.~35,
  pp.~35959--35970, 2022.

\bibitem{zhou2022non}
J.~Zhou, L.~Dong, Z.~Gan, L.~Wang, and F.~Wei, ``Non-contrastive learning meets
  language-image pre-training,'' {\em arXiv preprint arXiv:2210.09304}, 2022.

\bibitem{wang2024learn}
J.~Wang and G.~Kang, ``Learn to rectify the bias of clip for unsupervised
  semantic segmentation,'' in {\em Proceedings of the IEEE/CVF Conference on
  Computer Vision and Pattern Recognition}, pp.~4102--4112, 2024.

\bibitem{zhou2022learning}
K.~Zhou, J.~Yang, C.~C. Loy, and Z.~Liu, ``Learning to prompt for
  vision-language models,'' {\em International Journal of Computer Vision},
  vol.~130, no.~9, pp.~2337--2348, 2022.

\bibitem{zhou2022conditional}
K.~Zhou, J.~Yang, C.~C. Loy, and Z.~Liu, ``Conditional prompt learning for
  vision-language models,'' in {\em Proceedings of the IEEE/CVF Conference on
  Computer Vision and Pattern Recognition}, pp.~16816--16825, 2022.

\bibitem{he2022synthetic}
R.~He, S.~Sun, X.~Yu, C.~Xue, W.~Zhang, P.~Torr, S.~Bai, and X.~Qi, ``Is
  synthetic data from generative models ready for image recognition?,'' {\em
  arXiv preprint arXiv:2210.07574}, 2022.

\bibitem{zhang2022tip}
R.~Zhang, W.~Zhang, R.~Fang, P.~Gao, K.~Li, J.~Dai, Y.~Qiao, and H.~Li,
  ``Tip-adapter: Training-free adaption of clip for few-shot classification,''
  in {\em Computer Vision--ECCV 2022: 17th European Conference, Tel Aviv,
  Israel, October 23--27, 2022, Proceedings, Part XXXV}, pp.~493--510,
  Springer, 2022.

\bibitem{fei2006one}
L.~Fei-Fei, R.~Fergus, and P.~Perona, ``One-shot learning of object
  categories,'' {\em IEEE transactions on pattern analysis and machine
  intelligence}, vol.~28, no.~4, pp.~594--611, 2006.

\bibitem{lake2011one}
B.~Lake, R.~Salakhutdinov, J.~Gross, and J.~Tenenbaum, ``One shot learning of
  simple visual concepts,'' in {\em Proceedings of the annual meeting of the
  cognitive science society}, 2011.

\bibitem{krizhevsky2017imagenet}
A.~Krizhevsky, I.~Sutskever, and G.~E. Hinton, ``Imagenet classification with
  deep convolutional neural networks,'' {\em Communications of the ACM},
  vol.~60, no.~6, pp.~84--90, 2017.

\bibitem{he2016deep}
K.~He, X.~Zhang, S.~Ren, and J.~Sun, ``Deep residual learning for image
  recognition,'' in {\em Proceedings of the IEEE conference on computer vision
  and pattern recognition}, pp.~770--778, 2016.

\bibitem{simonyan2014very}
K.~Simonyan and A.~Zisserman, ``Very deep convolutional networks for
  large-scale image recognition,'' {\em arXiv preprint arXiv:1409.1556}, 2014.

\bibitem{finn2017model}
C.~Finn, P.~Abbeel, and S.~Levine, ``Model-agnostic meta-learning for fast
  adaptation of deep networks,'' in {\em International conference on machine
  learning}, pp.~1126--1135, PMLR, 2017.

\bibitem{ravi2017optimization}
S.~Ravi and H.~Larochelle, ``Optimization as a model for few-shot learning,''
  in {\em International conference on learning representations}, 2017.

\bibitem{bateni2020improved}
P.~Bateni, R.~Goyal, V.~Masrani, F.~Wood, and L.~Sigal, ``Improved few-shot
  visual classification,'' in {\em Proceedings of the IEEE/CVF Conference on
  Computer Vision and Pattern Recognition}, pp.~14493--14502, 2020.

\bibitem{snell2017prototypical}
J.~Snell, K.~Swersky, and R.~Zemel, ``Prototypical networks for few-shot
  learning,'' {\em Advances in neural information processing systems}, vol.~30,
  2017.

\bibitem{vinyals2016matching}
O.~Vinyals, C.~Blundell, T.~Lillicrap, D.~Wierstra, {\em et~al.}, ``Matching
  networks for one shot learning,'' {\em Advances in neural information
  processing systems}, vol.~29, 2016.

\bibitem{sung2018learning}
F.~Sung, Y.~Yang, L.~Zhang, T.~Xiang, P.~H. Torr, and T.~M. Hospedales,
  ``Learning to compare: Relation network for few-shot learning,'' in {\em
  Proceedings of the IEEE conference on computer vision and pattern
  recognition}, pp.~1199--1208, 2018.

\bibitem{hariharan2017low}
B.~Hariharan and R.~Girshick, ``Low-shot visual recognition by shrinking and
  hallucinating features,'' in {\em Proceedings of the IEEE international
  conference on computer vision}, pp.~3018--3027, 2017.

\bibitem{qi2018low}
H.~Qi, M.~Brown, and D.~G. Lowe, ``Low-shot learning with imprinted weights,''
  in {\em Proceedings of the IEEE conference on computer vision and pattern
  recognition}, pp.~5822--5830, 2018.

\bibitem{dhillon2019baseline}
G.~S. Dhillon, P.~Chaudhari, A.~Ravichandran, and S.~Soatto, ``A baseline for
  few-shot image classification,'' {\em arXiv preprint arXiv:1909.02729}, 2019.

\bibitem{joachims1999transductive}
T.~Joachims {\em et~al.}, ``Transductive inference for text classification
  using support vector machines,'' in {\em Icml}, vol.~99, pp.~200--209, 1999.

\bibitem{liu2018learning}
Y.~Liu, J.~Lee, M.~Park, S.~Kim, E.~Yang, S.~J. Hwang, and Y.~Yang, ``Learning
  to propagate labels: Transductive propagation network for few-shot
  learning,'' {\em arXiv preprint arXiv:1805.10002}, 2018.

\bibitem{bommasani2021opportunities}
R.~Bommasani, D.~A. Hudson, E.~Adeli, R.~Altman, S.~Arora, S.~von Arx, M.~S.
  Bernstein, J.~Bohg, A.~Bosselut, E.~Brunskill, {\em et~al.}, ``On the
  opportunities and risks of foundation models,'' {\em arXiv preprint
  arXiv:2108.07258}, 2021.

\bibitem{gao2021clip}
P.~Gao, S.~Geng, R.~Zhang, T.~Ma, R.~Fang, Y.~Zhang, H.~Li, and Y.~Qiao,
  ``Clip-adapter: Better vision-language models with feature adapters,'' {\em
  arXiv preprint arXiv:2110.04544}, 2021.

\bibitem{lu2022prompt}
Y.~Lu, J.~Liu, Y.~Zhang, Y.~Liu, and X.~Tian, ``Prompt distribution learning,''
  in {\em Proceedings of the IEEE/CVF Conference on Computer Vision and Pattern
  Recognition}, pp.~5206--5215, 2022.

\bibitem{xing2022class}
Y.~Xing, Q.~Wu, D.~Cheng, S.~Zhang, G.~Liang, and Y.~Zhang, ``Class-aware
  visual prompt tuning for vision-language pre-trained model,'' {\em arXiv
  preprint arXiv:2208.08340}, 2022.

\bibitem{zhu2022prompt}
B.~Zhu, Y.~Niu, Y.~Han, Y.~Wu, and H.~Zhang, ``Prompt-aligned gradient for
  prompt tuning,'' {\em arXiv preprint arXiv:2205.14865}, 2022.

\bibitem{sun2022dualcoop}
X.~Sun, P.~Hu, and K.~Saenko, ``Dualcoop: Fast adaptation to multi-label
  recognition with limited annotations,'' {\em arXiv preprint
  arXiv:2206.09541}, 2022.

\bibitem{deng2022rlprompt}
M.~Deng, J.~Wang, C.-P. Hsieh, Y.~Wang, H.~Guo, T.~Shu, M.~Song, E.~P. Xing,
  and Z.~Hu, ``Rlprompt: Optimizing discrete text prompts with reinforcement
  learning,'' {\em arXiv preprint arXiv:2205.12548}, 2022.

\bibitem{gao2020making}
T.~Gao, A.~Fisch, and D.~Chen, ``Making pre-trained language models better
  few-shot learners,'' {\em arXiv preprint arXiv:2012.15723}, 2020.

\bibitem{haviv2021bertese}
A.~Haviv, J.~Berant, and A.~Globerson, ``Bertese: Learning to speak to bert,''
  {\em arXiv preprint arXiv:2103.05327}, 2021.

\bibitem{jiang2020can}
Z.~Jiang, F.~F. Xu, J.~Araki, and G.~Neubig, ``How can we know what language
  models know?,'' {\em Transactions of the Association for Computational
  Linguistics}, vol.~8, pp.~423--438, 2020.

\bibitem{houlsby2019parameter}
N.~Houlsby, A.~Giurgiu, S.~Jastrzebski, B.~Morrone, Q.~De~Laroussilhe,
  A.~Gesmundo, M.~Attariyan, and S.~Gelly, ``Parameter-efficient transfer
  learning for nlp,'' in {\em International Conference on Machine Learning},
  pp.~2790--2799, PMLR, 2019.

\bibitem{jia2022visual}
M.~Jia, L.~Tang, B.-C. Chen, C.~Cardie, S.~Belongie, B.~Hariharan, and S.-N.
  Lim, ``Visual prompt tuning,'' in {\em Computer Vision--ECCV 2022: 17th
  European Conference, Tel Aviv, Israel, October 23--27, 2022, Proceedings,
  Part XXXIII}, pp.~709--727, Springer, 2022.

\bibitem{zhang2020side}
J.~O. Zhang, A.~Sax, A.~Zamir, L.~Guibas, and J.~Malik, ``Side-tuning: a
  baseline for network adaptation via additive side networks,'' in {\em
  Computer Vision--ECCV 2020: 16th European Conference, Glasgow, UK, August
  23--28, 2020, Proceedings, Part III 16}, pp.~698--714, Springer, 2020.

\bibitem{deng2009imagenet}
J.~Deng, W.~Dong, R.~Socher, L.-J. Li, K.~Li, and L.~Fei-Fei, ``Imagenet: A
  large-scale hierarchical image database,'' in {\em 2009 IEEE conference on
  computer vision and pattern recognition}, pp.~248--255, Ieee, 2009.

\bibitem{krause20133d}
J.~Krause, M.~Stark, J.~Deng, and L.~Fei-Fei, ``3d object representations for
  fine-grained categorization,'' in {\em Proceedings of the IEEE international
  conference on computer vision workshops}, pp.~554--561, 2013.

\bibitem{soomro2012ucf101}
K.~Soomro, A.~R. Zamir, and M.~Shah, ``Ucf101: A dataset of 101 human actions
  classes from videos in the wild,'' {\em arXiv preprint arXiv:1212.0402},
  2012.

\bibitem{fei2004learning}
L.~Fei-Fei, R.~Fergus, and P.~Perona, ``Learning generative visual models from
  few training examples: An incremental bayesian approach tested on 101 object
  categories,'' in {\em 2004 conference on computer vision and pattern
  recognition workshop}, pp.~178--178, IEEE, 2004.

\bibitem{nilsback2008automated}
M.-E. Nilsback and A.~Zisserman, ``Automated flower classification over a large
  number of classes,'' in {\em 2008 Sixth Indian Conference on Computer Vision,
  Graphics \& Image Processing}, pp.~722--729, IEEE, 2008.

\bibitem{xiao2010sun}
J.~Xiao, J.~Hays, K.~A. Ehinger, A.~Oliva, and A.~Torralba, ``Sun database:
  Large-scale scene recognition from abbey to zoo,'' in {\em 2010 IEEE computer
  society conference on computer vision and pattern recognition},
  pp.~3485--3492, IEEE, 2010.

\bibitem{cimpoi2014describing}
M.~Cimpoi, S.~Maji, I.~Kokkinos, S.~Mohamed, and A.~Vedaldi, ``Describing
  textures in the wild,'' in {\em Proceedings of the IEEE conference on
  computer vision and pattern recognition}, pp.~3606--3613, 2014.

\bibitem{helber2019eurosat}
P.~Helber, B.~Bischke, A.~Dengel, and D.~Borth, ``Eurosat: A novel dataset and
  deep learning benchmark for land use and land cover classification,'' {\em
  IEEE Journal of Selected Topics in Applied Earth Observations and Remote
  Sensing}, vol.~12, no.~7, pp.~2217--2226, 2019.

\bibitem{maji2013fine}
S.~Maji, E.~Rahtu, J.~Kannala, M.~Blaschko, and A.~Vedaldi, ``Fine-grained
  visual classification of aircraft,'' {\em arXiv preprint arXiv:1306.5151},
  2013.

\bibitem{parkhi2012cats}
O.~M. Parkhi, A.~Vedaldi, A.~Zisserman, and C.~Jawahar, ``Cats and dogs,'' in
  {\em 2012 IEEE conference on computer vision and pattern recognition},
  pp.~3498--3505, IEEE, 2012.

\bibitem{bossard2014food}
L.~Bossard, M.~Guillaumin, and L.~Van~Gool, ``Food-101--mining discriminative
  components with random forests,'' in {\em Computer Vision--ECCV 2014: 13th
  European Conference, Zurich, Switzerland, September 6-12, 2014, Proceedings,
  Part VI 13}, pp.~446--461, Springer, 2014.

\bibitem{kingma2014adam}
D.~P. Kingma and J.~Ba, ``Adam: A method for stochastic optimization,'' {\em
  arXiv preprint arXiv:1412.6980}, 2014.

\end{thebibliography}
}

\end{document}


\captionsetup{labelformat=empty} 

\begin{table*}
  \centering
  \begin{subtable}[b]{0.48\linewidth}
    \centering \captionsetup{labelformat=empty}
\begin{tabular}{lccccc}
\toprule 
\multicolumn{1}{c}{\multirow{2}{*}{\raisebox{-1ex}{Method}}} & \multicolumn{5}{c}{Number of shots} \\
\cmidrule(lr){2-6}
                        & 1     & 2     & 4    & 8    & 16   \\
\midrule
CoOp             & 67.6 & 70.3 & 73.6 & 76.4 & 79.3 \\
CoCoOp           & 68.7 & 70.4 & 71.7 & 73.7 & 75.7 \\
Tip-Adapter-F    & 71.3 & 73.4 & 76.1 & 78.9 & 81.1 \\
MaPLe            & 67.5 & 70.8 & 72.7 & 76.3 & 78.4 \\
MaPLe~(10~epoch) & 69.4 & 73.8 & 76.3 & 78.5 & 81.1 \\
MaPLe~(20~epoch) & 69.7 & 73.7 & 76.2 & 78.4 & 81.2 \\ 
\rowcolor[gray]{.9}
Ours             & \textbf{72.4} & \textbf{74.3} & \textbf{76.9} & \textbf{79.6} & \textbf{82.1} \\
\bottomrule
\end{tabular}
\vspace{-1mm}
    \caption{
    \fontsize{10}{7}\selectfont
    Table R1: Average accuracy of 11 datasets(ViT16).}
\vspace{-2mm}
    \label{subtableA}
  \end{subtable}
  \begin{subtable}[b]{0.48\linewidth}
    \centering \captionsetup{labelformat=empty}
\begin{tabular}{lccccc} 
\toprule[1pt]
\multicolumn{1}{c}{\multirow{2}{*}{\raisebox{-1ex}{Method}}} & \multicolumn{5}{c}{Number of shots} \\
\cmidrule(lr){2-6}
                        & 1     & 2     & 4    & 8    & 16   \\
\cmidrule(lr){1-6}
 Linear probe CLIP & 36.7 & 47.6 & 57.2 & 65.0 & 71.1 \\
              CoOp & 59.6 & 62.3 & 66.8 & 69.9 & 73.4 \\
      CLIP-Adapter & 62.7 & 65.5 & 68.6 & 71.3 & 74.4 \\
           ProGrad & 62.6 & 64.9 & 68.5 & 71.4 & 73.9 \\
       Tip-Adapter & 62.3 & 64.6 & 66.5 & 68.5 & 70.3 \\
     Tip-Adapter-F & 64.6 & 66.7 & 69.7 & 72.4 & 75.8 \\
              PLOT & 65.5 & \textbf{68.6} & \textbf{71.2} & 73.5 & 76.2 \\
\rowcolor[gray]{.9}
   Ours & \textbf{66.1} & 68.2 & 71.1 & \textbf{73.6} & \textbf{76.6} \\
\bottomrule[1pt]
\end{tabular}
\vspace{-1mm}
    \caption{
    \fontsize{10}{7}\selectfont
    Table R2: Average accuracy of 11 datasets~(ResNet50).}
\vspace{-2mm}
    \label{subtableB}
  \end{subtable}
\end{table*}


\begin{table*}
  \centering
  \begin{subtable}[b]{0.48\linewidth}
    \centering 
    \captionsetup{labelformat=empty}
\begin{tabular}{lcccc}
\toprule
Method       &Shots & Params & Params \%CLIP & Acc  \\ \midrule
CoOp         &1     & 2048   & 0.002         & 67.6 \\
MaPLe(cvpr23)&1     & 3.55M  & 2.85          & 69.7 \\
Tip-Adapter-F&1     & 0.82M  & 0.66          & 71.3 \\
RTOS         &1     & 3.94M  & 3.16          & 71.8 \\ \midrule
CoOp         &16    & 2048   & 0.002         & 79.3 \\
MaPLe(cvpr23)&16    & 3.55M  & 2.85          & 81.2 \\
Tip-Adapter-F&16    & 8.19M  & 6.58          & 81.1 \\
RTOS         &16    & 3.94M  & 3.16          & 81.3 \\ \bottomrule
\end{tabular}
\caption{
\fontsize{10}{7}\selectfont
Table R3: Comparison of computational complexity among different methods~(ViT-B/16). Acc denotes the average accuracy of 11 datasets.}
\vspace{-2mm}

    \label{subtableB}
  \end{subtable}
  \begin{subtable}[b]{0.48\linewidth}
    \centering \captionsetup{labelformat=empty}
\begin{tabular}{lccccc}
\toprule
Context Prompt     & 1 & 2 & 4 & 8 & 16 \\
\midrule
Unified~(learnable)           & 64.3 & 66.2 & 69.3 & 71.6 & 74.7 \\
Class-specific~(learnable)    & 64.1 & 66.1 & 69.0 & 71.1 & 74.7 \\
``A photo of [cls]"~(RTOS-C) & 65.0 & 67.0 & 70.0 & 72.4 & 75.4 \\
\bottomrule
\end{tabular}
\caption{
\fontsize{10}{7}\selectfont
Table R4: Average accuracy of 11 datasets~(ResNet50). Simultaneously training the context prompt in CoOp.}
\vspace{-2mm}
    \label{subtableC}
  \end{subtable}
\end{table*}

\begin{table}
\begin{center}
\begin{tabular}{llll}
\toprule
     & ZS-CLIP & CoOp & RTOS-C \\ \midrule
mIoU & 18.6           & 19.6 & 21.5   \\ \bottomrule
\end{tabular}
\end{center}
\vspace{-5mm}
\caption{Table R5: 
Semantic Segmentation Performance on VOC2012 with mIoU Metric~(ResNet50).
We evaluated the segmentation performance on the VOC2012 dataset using a subset of 65 images. This subset was carefully selected to ensure that each class is represented in approximately 4 images.
}
\vspace{-3mm}
\end{table}

\begin{table*}
\begin{center}
\begin{tabular}{lllllll}
\toprule
Methods        & Backbone & Training Cost & GFLOPs & Params  & Infer. Speed & GPU Memory \\ \midrule
Zero-shot CLIP   & RN50     & 0             & 6.14   & 39.34M  & 11.85ms      & 1648MiB    \\
CoOp             & RN50     & 880min        & 6.14   & 39.34M  & 11.85ms      & 1648MiB    \\
Synthetic        & RN50     & -             & 6.14   & 39.34M  & 11.85ms      & 1648MiB    \\
CLIP-Adapter     & RN50     & 50min         & 6.14   & 39.87M  & 12.00ms      & 1650MiB    \\
Tip-Adapter-F    & RN50     & 5min          & 6.17   & 71.72M  & 12.06ms      & 1776MiB    \\
PLOT             & RN50     & -             & 9.31   & 92.22M  & 29.59ms      & 2908MiB    \\
RTOS             & RN50     & 5min          & 6.32   & 53.10M  & 12.77ms      & 1702MiB    \\ \midrule
Zero-shot CLIP   & ViT-B16  & 0             & 17.58  & 86.70M  & 10.40ms      & 1710MiB    \\
CoOp             & ViT-B16  & 1724min       & 17.58  & 86.70M  & 10.40ms      & 1710MiB    \\
Synthetic        & ViT-B16  & -             & 17.58  & 86.70M  & 10.40ms      & 1710MiB    \\
CLIP-Adapter     & ViT-B16  & 73min         & 17.58  & 86.84M  & 10.66ms      & 1710MiB    \\
Tip-Adapter-F    & ViT-B16  & 7min          & 17.61  & 110.90M & 10.69ms      & 1804MiB    \\
MaPLe            & ViT-B16  & 66min         & 17.77  & 86.72M  & 11.08ms      & 1710MiB    \\
RTOS             & ViT-B16  & 8min          & 17.65  & 90.74M  & 11.23ms      & 1716MiB    \\ \bottomrule
\end{tabular}
\end{center}
\vspace{-6mm}
    \caption{
    Table R6: Comparison of time efficiency for different methods on 16-shot ImageNet.
    All experiments are tested with batch size 1 on a single NVIDIA Tesla V100 GPU.
    }
\label{tab:r2_major4_cpr_params_}
\vspace{-3mm}
\end{table*}

\begin{table*}
    \centering
\begin{center}
\begin{tabular}{lccccccccccc}
\toprule
\multicolumn{1}{c}{\multirow{2}{*}{\raisebox{-1ex}{Anchors}}} & \multicolumn{10}{c}{Target Dataset} \\ \cmidrule(lr){2-12}
 & \rotatebox{0}{ImageNet} & \rotatebox{0}{SUN.} & \rotatebox{0}{EuroSAT} & \rotatebox{0}{UCF.} & \rotatebox{0}{Food.} & \rotatebox{0}{Oxf.Pets} & \rotatebox{0}{Stan.Cars} & \rotatebox{0}{Flowers.} & \rotatebox{0}{FGVC.} & \rotatebox{0}{Caltech.} & \rotatebox{0}{DTD} \\
\midrule
ImageNet & 60.3 & 58.7 & 36.1 & 61.7 & 77.4 & 85.4 & 55.8 & 66.0 & 16.6 & 86.6 & 43.1 \\
SUN397 & 33.1 & 58.9 & 36.1 & 61.7 & 77.4 & 85.5 & 55.8 & 66.0 & 16.6 & 87.1 & 43.2 \\
EuroSAT & 60.3 & 58.9 & 36.2 & 61.9 & 77.4 & 85.8 & 55.8 & 66.1 & 17.0 & 85.9 & 43.0 \\
UCF101 & 59.6 & 58.8 & 36.2 & 61.9 & 77.4 & 85.5 & 55.8 & 66.0 & 16.6 & 86.5 & 42.9 \\
Food101 & 56.9 & 58.9 & 36.2 & 61.9 & 77.3 & 85.6 & 55.8 & 65.9 & 17.0 & 86.1 & 43.0 \\
OxfordPets & 60.3 & 58.8 & 37.2 & 61.9 & 77.4 & 85.8 & 55.7 & 66.0 & 17.0 & 85.9 & 43.2 \\
StandfordCars & 30.3 & 58.8 & 36.4 & 62.0 & 77.4 & 85.8 & 55.8 & 66.0 & 17.2 & 85.9 & 42.9 \\
Flowers102 & 46.2 & 58.9 & 36.2 & 61.8 & 77.4 & 85.8 & 55.8 & 66.1 & 16.8 & 86.1 & 42.9 \\
FGVCAircraft & 30.2 & 58.9 & 39.3 & 61.8 & 77.4 & 85.8 & 55.8 & 66.1 & 17.0 & 85.9 & 43.3 \\
Caltech101 & 60.3 & 58.8 & 37.2 & 61.8 & 77.4 & 85.5 & 55.8 & 66.1 & 16.5 & 85.8 & 43.0 \\
DTD & 60.3 & 58.8 & 37.7 & 61.8 & 77.4 & 85.5 & 55.8 & 66.1 & 16.7 & 87.5 & 42.8 \\
ImageNet21k & 30.2 & 58.6 & 35.9 & 60.3 & 77.4 & 83.3 & 55.7 & 65.9 & 16.7 & 86.8 & 43.1 \\
Zero-shot CLIP & 60.3 & 58.9 & 36.2 & 61.9 & 77.3 & 85.8 & 55.8 & 66.1 & 17.0 & 85.8 & 42.8 \\
\bottomrule
\end{tabular}
\end{center}
\vspace{-6mm}
    \caption{Table R7: Zero-shot results using 11 datasets as anchor datasets for each other.}
\vspace{-3mm}
\end{table*}
